\documentclass[10pt, a4paper]{article}

\usepackage[final]{lrec2026} 
\usepackage{subcaption}
\usepackage{svg}
\usepackage{booktabs}
\usepackage[title,titletoc]{appendix}
\usepackage{hyperref}
\usepackage{diagbox}
\usepackage{listings}
\usepackage{minted}
\usepackage[dvipsnames]{xcolor}
\usepackage{enumitem}

\definecolor{lightblue1}{RGB}{220, 235, 255}  
\definecolor{lightblue2}{RGB}{185, 210, 255}  
\definecolor{lightblue3}{RGB}{155, 195, 255}  

\definecolor{lightgreen1}{RGB}{205, 255, 205} 

\title{Contextualising (Im)plausible Events Triggers Figurative Language}

\name{Annerose Eichel, Tonmoy Rakshit, Sabine Schulte im Walde} 

\address{Institute for Natural Language Processing \\ 
        University of Stuttgart \\
         \texttt{\small {\{annerose.eichel,tonmoy.rakshit,schulte\}}@ims.uni-stuttgart.de}\\}

\abstract{
This work explores the connection between (non-)literalness and plausibility at the example of subject-verb-object events in English. We design a systematic setup of plausible and implausible event triples in combination with abstract and concrete constituent categories. Our analysis of human and LLM-generated judgments and example contexts reveals substantial differences 
between 
assessments of plausibility. 
While humans excel at nuanced detection and contextualization of (non-)literal vs. implausible events, LLM results reveal only shallow contextualization patterns with a bias to trade implausibility for non-literal, plausible interpretations.
 \\ \newline  }

\begin{document}

\maketitleabstract

\section{Introduction and Related Work} \label{sec:introduction}

How plausible do you judge a situation where the \textit{heat catches a cyclist}?
Based on the literal reading of this described situation, one might expect that the majority of human annotators would consider this situation as rather implausible. Yet, previous research has demonstrated that
humans are eager to interpret such a seemingly implausible event. In the current study, we investigate to which extent the interpretation of implausible events is driven by the potential to frame them in a non-literal context. For instance, the above situation has been contextualized in figurative example sentences 
such as \textit{During the intense desert race, the scorching heat caught the cyclist off guard, forcing him to stop for water and shade.} and \textit{The unexpected heat caught the cyclist unaware.} %

As the example illustrates, interpreting and distinguishing plausible from implausible events is a crucial and non-trivial building block of natural language. 
A range of work has explored semantic plausibility using 
subject-verb-object (svo) events in English leveraging embedding-based neural networks \cite{wang-etal-2018-modeling}, transformers \cite{porada-etal-2019-gorilla,emami-etal-2021-adept,porada-etal-2021-modeling}, and LLM-based methods \cite{kauf-etal-2024-log}.  However, prior research so far explicitly focused on literal events \cite{wang-etal-2018-modeling} or other aspects of plausibility. Our work addresses this gap and explores the connection between figurative language and plausibility.
To do so, we adopt definitions from previous work \cite{wilks1975,resnik1993,wang-etal-2018-modeling} and consider plausibility in a binary setting. \textit{Plausible} events include not only
highly typical events but also untypical events \cite{wilks1975}, potentially novel events \cite{wang-etal-2018-modeling}, and seemingly trivial events such as ``a person breathes'' that are not necessarily attested in an existing corpus \cite{gordon-vandurme-2013}. In comparison, fully implausible events do not allow any semantically valid interpretation; neither a literal nor a figurative reading, for example, through creative metaphors \cite{griciute-etal-2022-cusp}.

For our study, we rely on a small subset of svo event triples that were previously annotated as (im)plausible \cite{eichel-schulte-im-walde-2023-dataset}. Crucially, the original triples are balanced with regard to the degree of concreteness vs. abstractness of the involved constituent words, because concepts can be described in accordance with the way people perceive them \cite{barsalou2005situating,brysbaert2014}: Concrete concepts such as \textit{trampoline} can be seen, heard, touched, smelled, or tasted. In contrast, abstract concepts such as \textit{realism} cannot be perceived with the five senses. In between these two extremes on the scale, mid-range concepts  such as \textit{punctuality} are situated.
The provided abstractness information allows us to connect not only (im)plausibility to (non)literalness but to additionally integrate an interaction with conceptual abstractness, thus implicitly relating to Conceptual Metaphor Theory \cite{lakoff1980metaphors} as a mapping from abstract to concrete concepts to trigger metaphorical meanings as a special case of non-literal language.

More specifically, we make use of 411 svo triples with plausibility judgments, and ask humans and LLMs to make a binary judgement about their figurative language, plus providing example sentences. Our novel dataset contains a total of 6,497/14,555 judgments and 6,497/3,288 unique sentences generated by humans/LLMs.\footnote{\url{www.github.com/AnneroseEichel/NLE2026}} We use the collected judgements and sentences to compare human and LLM generations. In the context of plausibility, humans have been observed to tend towards sense-making with great nuance and willingness to interpret even whimsical sentences \cite{griciute-etal-2022-cusp,eichel-schulte-im-walde-2023-dataset}. Regarding LLMs, while they are equipped for semantic interpretation with (world) knowledge learned through distributional patterns in vast amounts of training data, almost all of their data are plausible. We thus formulate the following research questions:

\begin{itemize}[itemsep=1pt,leftmargin=6pt]
    \item RQ1: Does figurative language interact with event plausibility, and how does this interaction relate to the abstractness of the event constituents?
    \item RQ2: 
    How do human annotations compare to LLM judgments regarding figurative language and event plausibility?
    \item RQ3: Which qualitative differences can be observed for human vs. model-produced contextualizations of (non-)literal 
    implausible events?
\end{itemize}


Our contributions are threefold: (i) We present a collection of judgments and example contexts from humans and four LLMs, using a systematic setup of plausible and implausible event triples in combination with abstract and concrete constituent categories.
(ii) We provide detailed insights into human vs. LLM judgments for predicting (non-)literalness, and (iii) we conduct a careful analysis of human- and model-generated contexts as well as repair mechanisms for seemingly implausible events.   

\section{Data} \label{sec:data}
We use the plausibility dataset PAP \cite{eichel-schulte-im-walde-2023-dataset}. PAP encompasses a balanced set of 1,733 subject-verb-object triples in English extracted from Wikipedia (originally plausible) and automatically perturbed triples (originally implausible). All events are labeled by each component's concreteness ranging from abstract (a), over mid-range (m), to concrete (c) \cite{brysbaert2014}. PAP is balanced across all possible combinations of abstractness such as events consisting of only highly concrete words such as ``\textit{person calls town}'' (ccc) or fully mixed events such as ``\textit{career reestablishes chicken}'' (amc). 
Triples are annotated through crowd-sourcing with subjective assessments of plausibility on a degree scale (1--5) ranging from implausible to plausible. PAP ratings include raw annotations as well as original plausibility labels, and provide clear majority-based ($\geq 70\%$) aggregations. 
For this study, we use a subset satisfying the following criteria: across all abstractness combinations, we draw a random sample of event triples where (i) original and human-annotated majority label correspond to each other such as ``\textit{album breaks genre}'' (orig.: \textit{plausible}; PAP maj.: \textit{plausible}), and (ii) original and human-annotated majority label differ such as ``\textit{collection needs autonomy}'' (orig.: \textit{implausible}; PAP maj.: \textit{plausible}). An overview is shown in App.~\ref{app:data}, Table~\ref{tab:target-overview}.

\section{Methods}
For each svo event in our dataset sample, we collect judgments and example sentences from both humans and LLMs. We ask humans to (1) select a label for whether an event is figurative, literal, or neither, based on the combination of component meanings, and (2) produce an example contextualizing the event (only if figurative or literal), or a sentence contextualizing an altered event if neither (cf. App.~\ref{app:annotation}, Figure~\ref{fig:annotation-screenshot} for annotation instructions). 
Then, we prompt LLMs to complete the same tasks.
\paragraph{Human Annotation} \label{sec:annotation}



We use Prolific and Google Forms as study tools. Participants are required to 
reside in the UK, US, Ireland, or Australia and hold corresponding citizenship, speak English as their primary language, and have a Prolific approval rate of $\geq98\%$. Items are shown in batches of $\approx$25 items with one target shown per page. 
For each item, we collect $16^{\pm 1.45}$ responses from a total of 240 annotators. We make sure that each annotator contributes $<1\%$ to the collection. 
Our annotator sample has a median age of 38 years, is slightly skewed towards female (56.7\%) over male annotators (43.3\%), and resides mainly in the US (57\%). 
For full demographic details, cf. App.~\ref{app:annotation}.
Across abstractness combinations, we obtain a total of 6,497 judgments and example sentences.
\setlength{\abovecaptionskip}{2pt}
\setlength{\belowcaptionskip}{2pt}
\begin{figure*}[!htbp]
    \centering
    \begin{subfigure}[b]{0.5\textwidth}
         \centering
        \includegraphics[width=\linewidth]{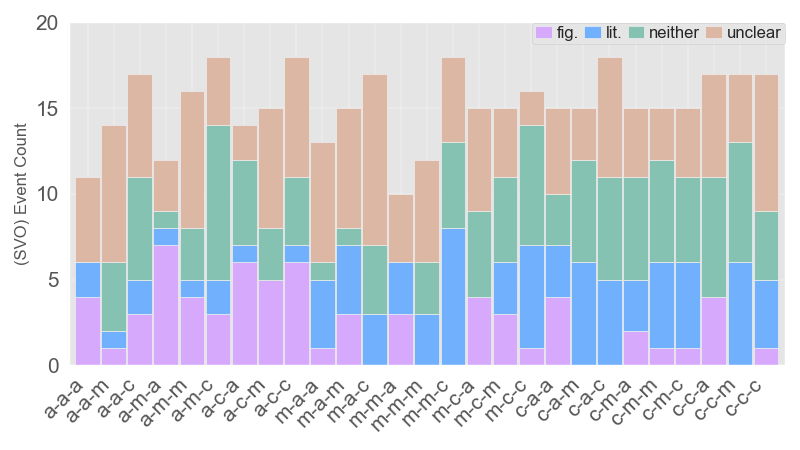}
        \caption{\textbf{Human} judgements across abstractness comb.}
    \end{subfigure}%
         \begin{subfigure}[b]{0.225\textwidth}
         \centering
        \includegraphics[width=\linewidth]{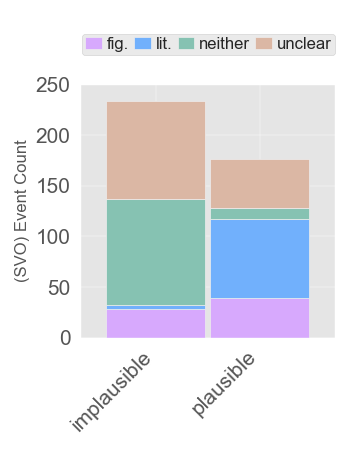}
        \caption{orig. label}
    \end{subfigure}%
    \begin{subfigure}[b]{0.225\textwidth}
        \centering
        \includegraphics[width=\linewidth]{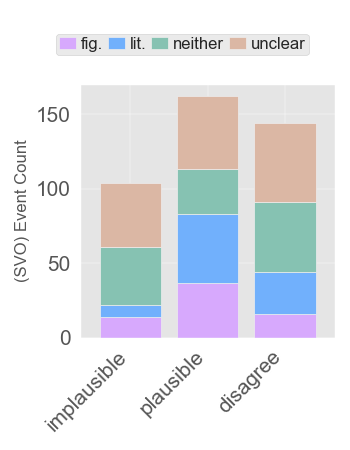}
        \caption{PAP rating}
    \end{subfigure}
    \hfill
    \caption{Analysis of \textbf{human} figurative majority-assigned labels assigned across (a) 27 \textbf{abstractness combinations} ranging from most abstract on the left to most concrete one the right, (b) in comparison with \textbf{original labels} used to create PAP, and (c) \textbf{PAP majority ratings}. }
    \label{fig:human-label-analysis}
\end{figure*}
\vspace{-1.7em}

\setlength{\abovecaptionskip}{2pt}
\setlength{\belowcaptionskip}{2pt}
\begin{figure*}[!htbp]
    \centering
    \begin{subfigure}[b]{0.5\textwidth}
         \centering
        \includegraphics[width=\linewidth]{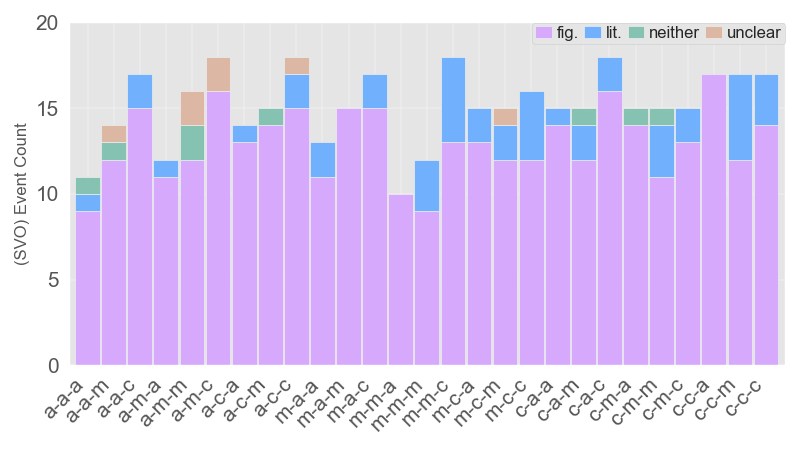}
        \caption{\textbf{\texttt{Qwen3-4B}} judgements across abstractness comb.}
    \end{subfigure}%
         \begin{subfigure}[b]{0.225\textwidth}
         \centering
        \includegraphics[width=\linewidth]{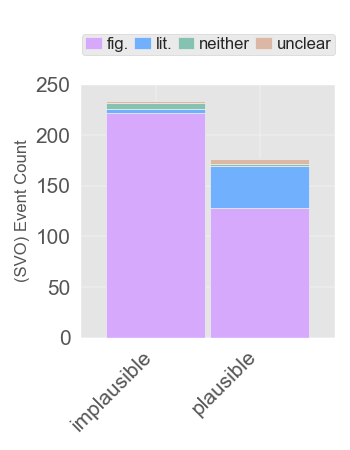}
        \caption{orig. label}
    \end{subfigure}%
    \begin{subfigure}[b]{0.225\textwidth}
        \centering
        \includegraphics[width=\linewidth]{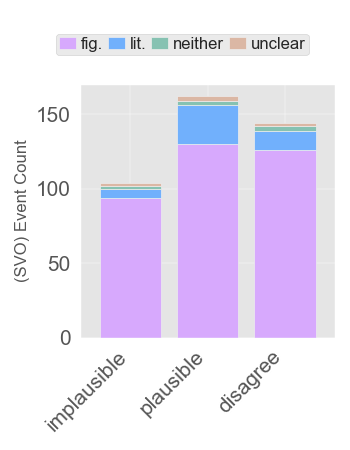}
        \caption{PAP rating}
    \end{subfigure}
    \hfill
    \caption{Analysis of \textbf{\texttt{Qwen3-4B}} 
    figurative labels (zero-shot), similarly to Figure~\ref{fig:human-label-analysis}.}
    \label{fig:model-label-analysis}
    \vspace{-1.6em}
\end{figure*}

\paragraph{Modeling} \label{sec:modeling}
For model predictions, we use the Instruct versions of four 
LLMs. We focus on moderate parameter count and test four multilingual model families:  Qwen3-4B \cite{qwen3technicalreport}, Gemma3-4B \cite{gemma_2025}, Mistral-7B \cite{jiang2023mistral7b}, and Llama3.1-8B \cite{grattafiori2024llama3herdmodels}. 
The LLM prompts for label and text generation are based on instructions for humans with (few-shot) and without (zero-shot) examples (cf. App.~\ref{subsec:modeling}, Figure~\ref{fig:prompts-zero-shot} and Figure~\ref{fig:prompts-few-shot} for prompts).
(1) For label prediction, we aggregate results across five model runs with different random seeds and three prompts with varying phrasing and output formats. (2) For example sentence generation, we replicate human instructions as closely as possible and obtain generations for one seed. 
All prompting is performed with model default settings and a 64 token limit. 

\section{Results} 

\subsection{Figuratively, literally, unclear, or actually not plausible at all?}

To explore RQ1, i.e., how figurative language interacts with event plausibility and event constituent abstractness, we first provide a deep-dive into human judgments. To shed light on RQ2, we then assess how human annotations compare to SOTA LLMs' judgments.

\paragraph{Human Judgments} \label{par:humanjd} 

To investigate the relationship between \textbf{abstractness} as well as original and PAP \textbf{plausibility} ratings and figurative meanings of the targets, we visualize the number of svo events judged by the majority of participants. Here, majority is defined as $\geq60\%$. Whenever no majority is reached, we assign the label \textit{unclear}. Figure~\ref{fig:human-label-analysis} presents three perspectives on the distribution of judgments. Across plots, label categories are: \textit{figurative} (violet), \textit{literal} (blue), \textit{neither} (no contextualization is possible, green), and \textit{unclear} (orange).

We first look at the interplay between the \textbf{abstractness of the svo events, and their perception as figurative vs. literal language}. 
Across the $x$-axis, we observe a clear trend. More concrete events (e.g., (\textit{ccc}), (\textit{cam}), (\textit{cac})) are judged more literally, while more abstract events (e.g., (\textit{aaa}), (\textit{ama}), (\textit{mca})) are judged more figuratively. More specifically, subject and object abstractness exert greater influence on (non-)literalness. In particular, concrete or mid-scale verbs in such events lead to predominantly literal readings of events, confirming prior work \cite{khaliq-etal-2024-comparison, knuples-etal-2026-lit}.
In turn, we observe a limited influence of verb abstractness.
Items for which neither a figurative nor a literal reading or a context could be inferred such as ``\textit{payload lives blowout}'' or ``\textit{city folds fruit}'' accumulate on the more concrete end of the scale. Here, concrete objects strongly influence events that are otherwise abstract. 

Next, we focus on the relationship between \textbf{plausibility and  figurative language}. We consider both the original labels underlying the PAP dataset, and the  majority PAP ratings. 
Originally plausible items such as ``\textit{advance guarantees freedom}''
are judged more figuratively than originally implausible items such as ``\textit{copy inflates disbelief}''. While the difference is rather small for original labels, a larger disparity is observed for PAP ratings.
When inspecting targets judged literally such as ``\textit{owner secures trademark}'', a virtual clear-cut between plausible and implausible items emerges. While both original and majority PAP plausible items are judged mainly literally, implausible items are rarely perceived as literal. 

\begin{table*}[!htpb]
\centering
\small
\begin{tabular}{@{}l|r|rr|rr|rr|rr|rr@{}}
\toprule

\multicolumn{2}{l}{\hspace*{-2mm}\textbf{}}
& \multicolumn{2}{l}{\textbf{Human}} & \multicolumn{2}{l}{\textbf{Qwen3-4B}} & \multicolumn{2}{l}{\textbf{Llama3.1-8B}} & \multicolumn{2}{l}{\textbf{Mistral-7B}} & \multicolumn{2}{l}{\textbf{Gemma3-4B}} \\ \midrule
\textbf{Figurative}  &  df & $\chi^{2}$    & $V$ & $\chi^{2}$    & $V$ & $\chi^{2}$    & $V$ & $\chi^{2}$    & $V$ & $\chi^{2}$    & $V$           \\
\midrule
\textsc{abstractness} & 26 & \textbf{63.64***}  & 0.39 & 23.10                  & 0.24                & 48.13** & 0.34 & 41.50*   & 0.32 & 42.82*            & 0.32           \\
\textsc{orig. label}       & 1  & 6.91**    & 0.13 & 37.68                 & 0.30                 & 0.17    & 0.02 & 4.64*   & 0.11 & 33.10              & 0.28           \\
\textsc{PAP rating} & 3  & 13.49**   & 0.18 & 6.20                   & 0.12                & 2.88    & 0.08 & 1.59    & 0.06 & 7.68              & 0.14           \\
\midrule
\multicolumn{12}{l}{\hspace*{-2mm}\textbf{Literal}}\\
\midrule
\textsc{abstractness} & 26 & 43.93**   & 0.33 & 34.81                 & 0.29                & 50.88** & 0.35 & 42.67*  & 0.32 & 43.75*            & 0.33           \\
\textsc{orig. label}       & 1  & \textbf{111.33***} & 0.52 & 45.72                 & 0.33                & 0.46    & 0.03 & 4.66*   & 0.11 & 32.23             & 0.28           \\
\textsc{PAP rating} & 3  & \textbf{17.29***}  & 0.21 & 7.85*                 & 0.14                & 2.25    & 0.07 & 5.42    & 0.11 & 8.09*             & 0.14           \\
 \midrule

\multicolumn{12}{l}{\hspace*{-2mm}\textbf{Neither}}\\
\midrule
\textsc{abstractness} & 26 & 32.80      & 0.28 & 27.33                 & 0.26                & 23.33   & 0.24 & -       & -  & 32.86             & 0.28           \\
\textsc{orig. label}       & 1  & 71.96     & 0.42 & 0.45                  & 0.03                & 1.53    & 0.06 & -       & -  & 1.19              & 0.05           \\
\textsc{PAP rating} & 3  & 13.73**   & 0.18 & 0.04                  & 0.01                & 0.48    & 0.03 & -       & -  & 2.16              & 0.07           \\ \midrule
\multicolumn{12}{l}{\hspace*{-2mm}\textbf{Unclear}}\\
\midrule
\textsc{abstractness} & 26 & 29.59     & 0.27 & 32.56                 & 0.28                & 28.31   & 0.26 & 39.61*  & 0.31 & 40.18*            & 0.31           \\
\textsc{orig. label}       & 1  & 8.23**    & 0.14 & 1.33                  & 0.06                & 0.80     & 0.04 & 0.32    & 0.03 & 1.50               & 0.06           \\
\textsc{PAP rating} & 3  & 4.17      & 0.10  & 0.15                  & 0.02                & 2.10     & 0.07 & 0.59    & 0.04 & 5.42              & 0.11           \\ \bottomrule

\end{tabular}
\caption{Associations between figurative language and abstractness, original label, or PAP ratings. $\chi^{2}$ indicates \textit{significance} ($p<.05$:*, $p<.01$:**, $p<.001$:***) and Cramér's $V$ measures \textit{strength} of association. Model results are based on zero-shot prompts.} 
\label{tab:label-specific-chi-cramersv-all}
\vspace{-1.7em}
\end{table*}

Following our qualitative inspection, we further \textbf{quantify whether assigned (non-)literal labels are related with event abstractness, original labels, or PAP ratings} using $\chi^{2}$ tests of independence \cite{pearson1991correlated}. Strength of association is examined with Cramér's $V$ \cite{cramer1999mathematical}. 
An overview of results is presented in Table~\ref{tab:label-specific-chi-cramersv-all}. We find significant 
associations ($p<.001$) between abstractness combinations and figurative/literal labels, with a moderate effect size. 
This finding further underlines previous work \cite{khaliq-etal-2024-comparison, knuples-etal-2026-lit} focusing on verb-object pairs 
where they find an increase in figurative majority-based judgments predominantly influenced by object abstractness.
We further find a significant association ($p<.001$) between original and literal labels, with a strong effect size. For literal labels and majority PAP, we also observe a statistically reliable association ($p<.001$) albeit with a weaker effect size. 

\paragraph{Human vs. LLM Judgments} We compare majority-assigned label distributions obtained by humans vs. four SOTA LLMs where majority is defined as $\geq60\%$. 
Results across models and prompt types are shown in Table~\ref{tab:label-prediction} with performance equally low for all four models, i.e., they mostly disagree with human judgments. When visualizing majority-assigned labels, 
we observe clear differences across models and prompt types. For reasons of space, we illustrate model results at the example of Qwen3-4B results in Figure~\ref{fig:model-label-analysis} and provide a full overview in App.~\ref{app:results}, Figure~\ref{fig:full-model-label-analysis}.

\textbf{Zero-shot prompts} trigger Gemma, Qwen, and Mistral to assign overwhelming majorities of plausible, and specifically figurative readings. This holds for both originally plausible and implausible events. 
A notable exception is Llama which assigns significantly more literal interpretations across original labels and abstractness combinations. This trend is only partially observable for the other three models which assign literal readings for more concrete events.
In comparison to the label distribution based on human annotations, there is a notable absence of implausible instances across all models. In particular, Mistral does not produce a single majority assignment for \textit{implausible}. 
Similarly to the analysis of human judgements, we conduct a quantitative analysis to explore the relationship between figurative language and abstractness, original label, and PAP rating. Results are shown in Table~\ref{tab:label-specific-chi-cramersv-all}, indicating associations ($p<[0.01, 0.05]$) with moderate effect size between \textit{figurative} and \textit{literal} labels and abstractness for all models except Qwen. 

\textbf{Few-shot prompts} (cf. App.~\ref{app:results}, Figure~\ref{fig:full-model-label-analysis-fewshot}) change Gemma results with a significant increase in implausible and unclear events. 
Interestingly, only marginally different results are observable for both Qwen and Mistral, which could either point to strong prediction stability across prompts or disregard of contextual information in the middle of a prompt. Lastly, Llama results change to overall more figurative events assigned. Additional quantitative inspections (cf. App.~\ref{app:results}, Table~\ref{tab:label-specific-chi-cramersv-all-fewshot}) underline the relation between plausible labels and abstractness with stronger associations than for zero-shot prompting. 

We further explore for both zero- and few-shot settings \textbf{which prompt template leads to the strongest bias towards figurative interpretation} of the examined events. Results are shown in App.~\ref{app:results}, Table~\ref{tab:prompt-analysis}, indicating that prompt templates based on human instructions introduce the least bias across models.

\smallskip
\begin{table}[!thpb]
\centering
\small
\begin{tabular}{@{}lrrrr@{}}
\toprule
\multicolumn{1}{c}{\textsc{model}} & \multicolumn{2}{c}{\textsc{zero-shot}}      & \multicolumn{2}{c}{\textsc{few-shot}}    \\ \cmidrule{2-5}
\multicolumn{1}{c}{}      & \multicolumn{1}{c}{\textsc{acc.}}     & \multicolumn{1}{c}{$\tau$} & \multicolumn{1}{c}{\textsc{acc.}}        & \multicolumn{1}{c}{$\tau$} \\ \midrule
Gemma3-4B                 & 0.27                        & 0.09                        & 0.30                           & 0.01                        \\
Qwen3-4B                  & 0.27                        & \textbf{-0.15}                       & 0.25                           & \textbf{-0.16}                   \\
Mistral-7B                & 0.20                        & -0.10                       & 0.21                           & -0.05                       \\
Llama3.1-8B               & 0.28                        & 0.04                        & 0.32                           & 0.05                        \\ \bottomrule
\end{tabular}
\caption{Model performance for label prediction across prompt types. Predictions are aggregated across prompt templates and five model runs. We report \textit{accuracy} (acc.) and \textit{Kendall's} $\tau$ using human majority-assigned decisions as reference value. Bold: $p<0.001$}
\label{tab:label-prediction}
\vspace{-0.6cm}
\end{table}
In summary, our hypotheses are confirmed for \textbf{human-annotated} events: (i) The more concrete svo event constituents are, the more likely contextualization fails, i. e., events being judged as neither figuratively nor literally meaningful, but implausible (nonsensical). 
This finding underlines previous work on the influence of event abstractness on plausibility \cite{eichel-schulte-im-walde-2023-dataset} and complements research on semantically anomalous vs. truly nonsensical expressions \cite{olsen2026finding}.  
(ii) Confirming our hypothesis, the more abstract svo event constituents are, the more frequently plausibility is perceived, and the more probable is a figurative reading. 
In contrast, \textbf{LLM-predicted} results deviate from our hypothesis as we find (i+ii) a strong bias for plausibility, and specifically figurative language across categories for Qwen, Mistral and Gemma, while overall, Llama results are closer to human judgments. However, 
human-annotated results are only weakly mirrored with models trading implausibility for plausibility. 

\subsection{Qualitative Characteristics of Human- vs. LLM-Produced Contexts}
We qualitatively evaluate generated contexts to assess how humans vs. models contextualize (im)plausible svo events. Across our 27 abstractness combinations, we sample up to four examples (one per label) produced by humans or models (zero-shot)
, yielding 97 contexts.
We sample 3 example sentences per investigated event from human-generated contexts. We label one model generation per event. Events incorrectly (not) containing the original svo event are labeled \textit{none}. We also assign \textit{none} in case of more than one changed constituent or in case of constituent changes despite a plausible judgement.
Whenever events correctly contain the original event but are semantically invalid, we assign the label \textit{anomalous}. We follow \citet{olsen2026finding}'s labeling scheme and annotate contexts as \textit{specific} if no self-reported indication of generic settings such as \textit{fantasy} story is present in produced example contexts. In case of changes, we track altered event constituents. 

Results for human-produced and model-generated contexts are reported in Table~\ref{tab:examples-sentences}, highlighting a substantial number of specific contexts by both humans and LLMs. 
\begin{table}[!htpb]
\centering
\small
\begin{tabular}{@{}l|rrr|rrrr@{}}
\toprule
                      \multicolumn{1}{r}{}  & \multicolumn{1}{r}{H1}      & H2 & \multicolumn{1}{r}{H3} & \multicolumn{1}{r}{LL} & QW & MI. & GE\\ \midrule
Specific                & 94           & 92      & 93      & 51          & 52 & 27 & 17         \\ \midrule
Altered (s)          & 10            & 8       & 8       &     -        &        -   &     -        &        2    \\
Altered (v)          & 6            & 13      & 13      &     -        &       -     &     -        &        -    \\
Altered (o)           & 15           & 8       & 8       &    -         &         -  &     -        &        -   \\ \midrule
Anomalous               &    -          & 1       & 2       & 6           & 11    & 1 & 22      \\
None                    & 3            & 4       & 2       & 40          & 34 & 69 & 58         \\
\bottomrule
\end{tabular}
\caption{Human (H) vs. LLM (LL: Llama, QW: Qwen, MI: Mistral, GE: Gemma) context patterns.}
\label{tab:examples-sentences}
\vspace{-0.6cm}
\end{table}
In comparison to humans, LLMs rarely predicted the label \textit{neither} in which case an event constituent should be altered to enable contextualization. Nevertheless, especially Mistral and Gemma frequently alter events despite a predicted figurative or literal label.
Moreover, in the few cases were \textit{neither} was assigned, models mostly fail to correctly change only one constituent \textit{and} produce a valid sentence. Further, LLM contextualization strongly adheres to original event syntax, as highlighted by contexts to the originally and majority PAP implausible event \textit{``license hinders ice''}. Qwen repeats the event (\textit{``The event license hinders ice.''}). Both Llama and Gemma add a single object (\textit{``The event license hinders ice skating.''}). Mistral's generation illustrates a common failure across all models: incorrect substitution despite a plausible judgement 
(\textit{``The ice sculpture exhibition was hindered by the event license restrictions.'')}

In comparison, human-produced contexts are based on a majority-assigned \textit{neither} label with examples altering the subject (\textit{``The recent sunny weather hinders ice for hockey player.''}) or the object (\textit{``The strict license hindered access to the restricted research facility.''}) or providing an actually supporting, non-anomalous context (\textit{``A license doesn't have the capability to hinder ice from forming.''})  While humans use a wide vocabulary range and vary syntax to generate meaningful contexts, all models over-use the term \textit{event}, 
and adhere to mostly nouns, verbs, and simple syntactic structures. 
In conclusion, LLM results reveal shallow patterns when compared with human-generated contexts exhibiting great nuance at assessing plausibility and contextualizing and `repairing' events.



\section{Conclusion} \label{sec:conclusion}
This work explored the connection between plausibility and (non-)literalness at the example of svo events in English. 
Using a carefully selected section of the PAP dataset \cite{eichel-schulte-im-walde-2023-dataset}, we collected and analyzed human- vs. LLM-generated judgments and examples. 
Our analysis reveals substantial differences between human and LLM 
assessments for the examined events. While humans excel at nuanced detection and contextualization of (non-)literal vs. implausible events, LLM results reveal shallow context patterns and a strong bias towards plausibility.





\newpage
  
\section{Acknowledgements}
This research was supported by the Hanns Seidel Foundation's Talent Program (first author) and the DFG Research Grant SCHU 2580/4 \textit{MUDCAT -- Multimodal Dimensions and Computational Applications of Abstractness}. We also thank the reviewers for their helpful comments and nuanced feedback.

\section{Limitations and Ethical Considerations}

A first obvious limitation of our work is the sole focus on the English language. We expect results to differ for other languages and encourage work on plausibility, (non-)literalness, and abstractness.
In this paper, we present a collection of (non-)literalness judgments and examples sentences collected via crowd-sourcing. We employ control items as well as post-processing to minimize the impact of unreliable annotations on our analyses. Approaches of mitigation could be concentrating on labels with high majorities of one label assigned or use e. g., probabilistic approaches to aggregate labels. 
We pay participants fairly and seek transparent communication of decisions whenever necessary during the annotation approval process. 
Furthermore, in our work we use a reasonable set of heuristics to parse LLM-generations. It is possible that more complex approaches might lead to different results based on the parsing process.
Finally, we use LLMs that are known to exhibit bias which might be reflected in the way events are judged as well as in the style and content of generated contexts. 








\section{Bibliographical References}\label{sec:reference}

\bibliographystyle{lrec2026-natbib}
\bibliography{lrec2026-example}


\begin{appendices}
\section{Appendix}

\subsection{Data} \label{app:data}

Table~\ref{tab:target-overview} presents an overview of the target svo events sampled from the PAP dataset.

\begin{table}[!htbp]
\centering
\small
\begin{tabular}{@{}lrrr@{}}
\toprule
                       \textbf{orig. labels}    & \multicolumn{3}{c}{\textbf{PAP ratings}}                                                                                           \\ 
                       \cmidrule{2-4}
                           & \multicolumn{1}{c}{\textit{plausible}} & \multicolumn{1}{c}{\textit{disagree}} & \multicolumn{1}{c}{\textit{implausible}} \\ \midrule
\textit{plausible}  & 81                                     & 64                                    & 31                                       \\
\textit{implausible} & 81                                     & 80                                    & 77                                       \\ \bottomrule
\end{tabular}
\caption{Overview of number of target event triples sampled from PAP.}
\label{tab:target-overview}
\end{table}

\subsection{Human Annotation} \label{app:annotation}

\begin{figure*} 
    \centering
    \includegraphics[width=1\linewidth]{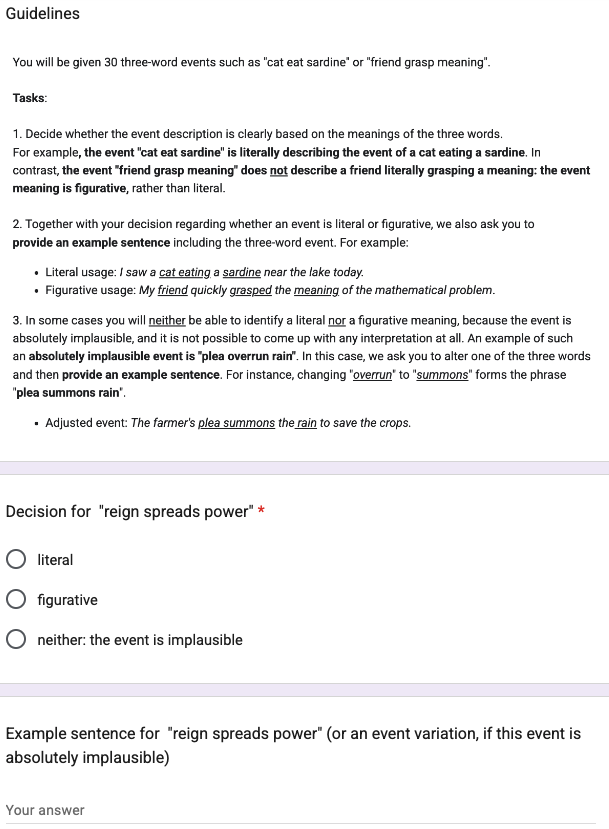}
    \caption{Annotation interface}
    \label{fig:annotation-screenshot}
\end{figure*}

\paragraph{Human Annotator Demographics}

\begin{figure*}[!htbp]
    \centering
    \begin{subfigure}[b]{0.47\textwidth}
         \centering
        \includegraphics[width=\linewidth]{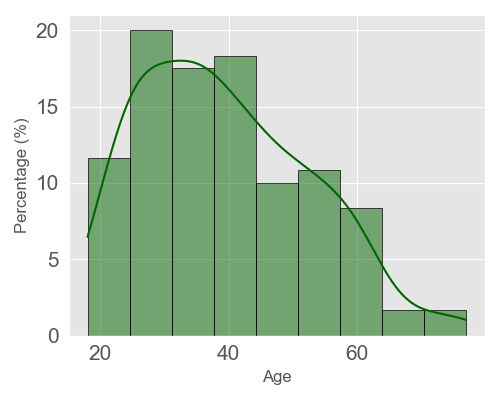}
        \caption{Age.}
    \end{subfigure}%
    \hfill
         \begin{subfigure}[b]{0.445\textwidth}
         \centering
        \includegraphics[width=\linewidth]{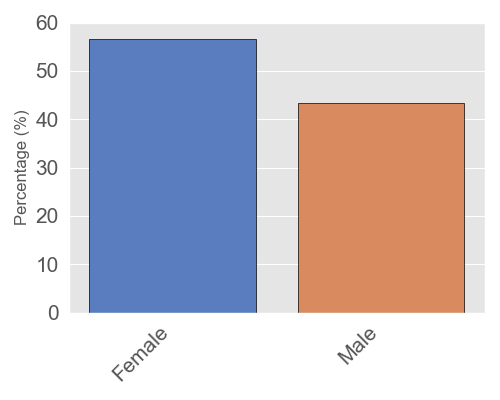}
        \caption{Female and male (sex; gender identity information not collected).}
    \end{subfigure}%
    \hfill
    \begin{subfigure}[b]{0.505\textwidth}
        \centering
        \includegraphics[width=\linewidth]{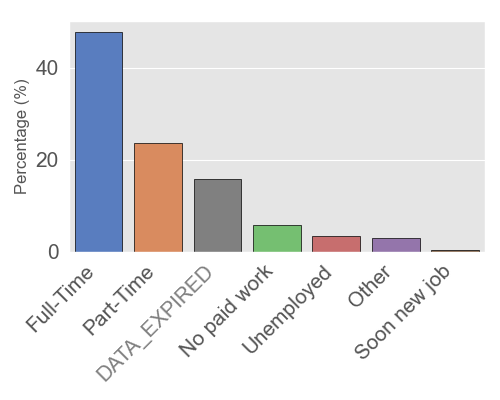}
        \caption{Employment status.}
    \end{subfigure}%
    \hfill
    \begin{subfigure}[b]{0.495\textwidth}
        \centering
        \includegraphics[width=\linewidth]{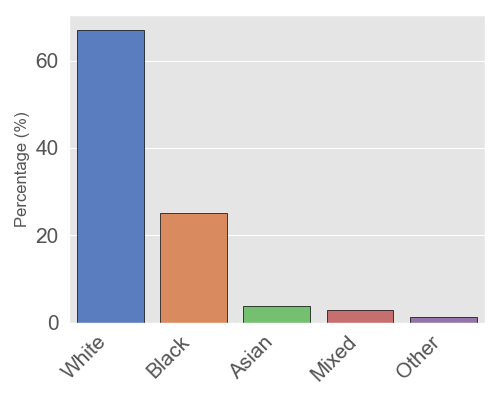}
        \caption{Self-reported simplified ethnicity groups.}
    \end{subfigure}%
            
    \hfill
    
    \begin{subfigure}[b]{0.5\textwidth}
        \centering
        \includegraphics[width=\linewidth]{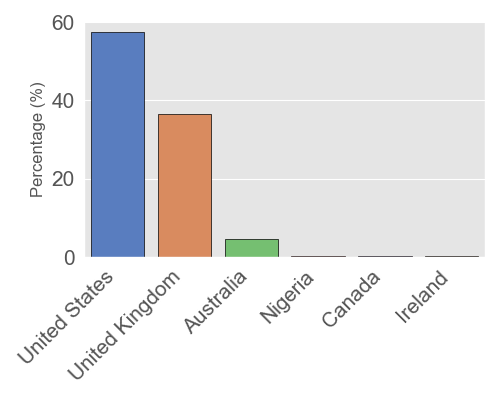}
        \caption{Nationalities (dual citizenship possible).}
    \end{subfigure}%
        \begin{subfigure}[b]{0.5\textwidth}
        \centering
        \includegraphics[width=\linewidth]{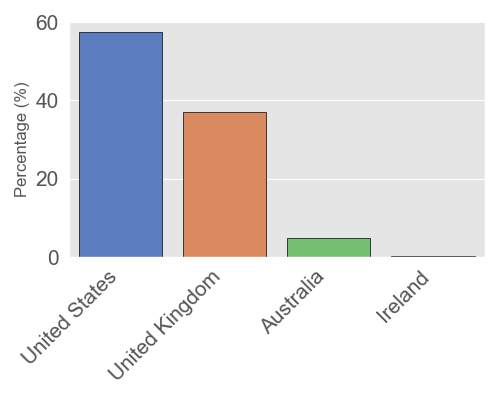}
        \caption{Country of residence.}
    \end{subfigure}
    \hfill
    \caption{Distributions of annotator demographic features.}
    \label{fig:anno}
    \vspace{-0.5cm}
\end{figure*}

Figure~\ref{fig:anno} shows an overview of annotator demographics. Subplot (a) visualizes the age distribution of involved annotators. Mean age is 39.5 years and median age is 38 years. Subplot (b) presents the distribution between female and male participants. Please note that this is based on self-reported biological sex of participants. We do not collect information on gender identity. We report annotators' employment status in subplot (c) with the majority of participants either working full-time or part-time. ``No paid work'' refers to individuals focusing on care work as well as retired or disabled individuals. ``Soon new job'' denotes participants who start a new job in the next month (which does not mean that they are not employed at the moment they took part in the study). ``Unemployed'' implies that someone is unemployed \textit{and} job-seeking. ``DATA\_EXPIRED'' refers to long-time participants on Prolific who have not updated their Prolific profile for a longer period of time. Some information such as employment or student status hence might get marked as expired. As visualized in subplot (d), self-reported simplified ethnicity groups are mainly White (67\%) and Black (25\%). While nationality as shown in subplot (e) needs to include UK, Ireland, U.S., or Australia, participants might have dual citizenship (e.g., the UK allows for that). Subplot (f) lists countries where participants reside. 

\setlength{\abovecaptionskip}{2pt}
\setlength{\belowcaptionskip}{2pt}

\subsection{Modeling} \label{subsec:modeling}

We present prompt templates for zero- and few-shot prompting in Figure~\ref{fig:prompts-zero-shot} and Figure~\ref{fig:prompts-zero-shot}.

\section{Results} \label{app:results}



\paragraph{Human vs. Model Label Predictions} 

\begin{figure*}[!htbp]
    \centering
        \setcounter{subfigure}{0}
    \renewcommand{\thesubfigure}{1\alph{subfigure}}
    \begin{subfigure}[b]{0.5\textwidth}
         \centering
        \includegraphics[width=\linewidth]{plots/Qwen3-4B-Instruct-2507--ZS-overall_color_decision_plot.png}
        \caption{\textbf{\texttt{Qwen3-4B}} judgements across abstractness comb.}
    \end{subfigure}%
         \begin{subfigure}[b]{0.225\textwidth}
         \centering
        \includegraphics[width=\linewidth]{plots//Qwen3-4B-Instruct-2507-orig-vs-literalness.png}
        \caption{orig. label}
    \end{subfigure}%
    \begin{subfigure}[b]{0.225\textwidth}
        \centering
        \includegraphics[width=\linewidth]{plots/Qwen3-4B-Instruct-2507-literalness-vs-pap-ratings-plot.png}
        \caption{PAP rating}
    \end{subfigure}
    \hfill
    \setcounter{subfigure}{0}
    \renewcommand{\thesubfigure}{2\alph{subfigure}}
    \begin{subfigure}[b]{0.5\textwidth}
         \centering
        \includegraphics[width=\linewidth]{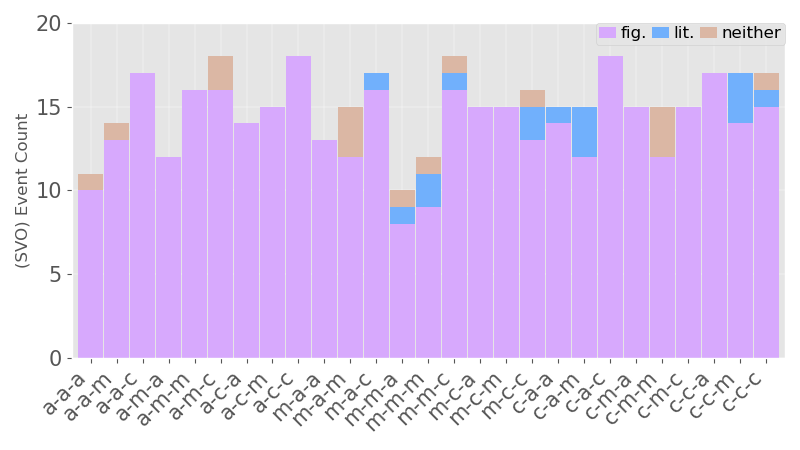}
        \caption{\textbf{\texttt{Mistral-7B}} judgements across abstractness comb.}
    \end{subfigure}%
         \begin{subfigure}[b]{0.225\textwidth}
         \centering
        \includegraphics[width=\linewidth]{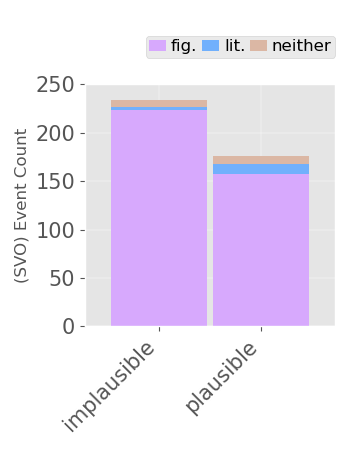}
        \caption{orig. label}
    \end{subfigure}%
    \begin{subfigure}[b]{0.225\textwidth}
        \centering
        \includegraphics[width=\linewidth]{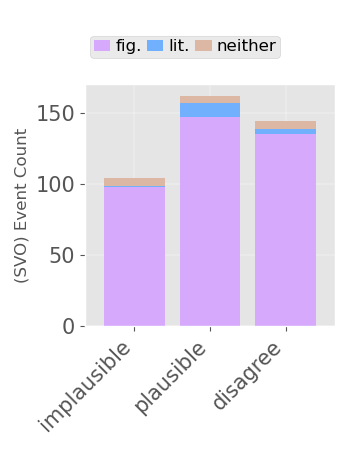}
        \caption{PAP rating}
    \end{subfigure}
    \hfill
\setcounter{subfigure}{0}
\renewcommand{\thesubfigure}{3\alph{subfigure}}
    \begin{subfigure}[b]{0.5\textwidth}
         \centering
        \includegraphics[width=\linewidth]{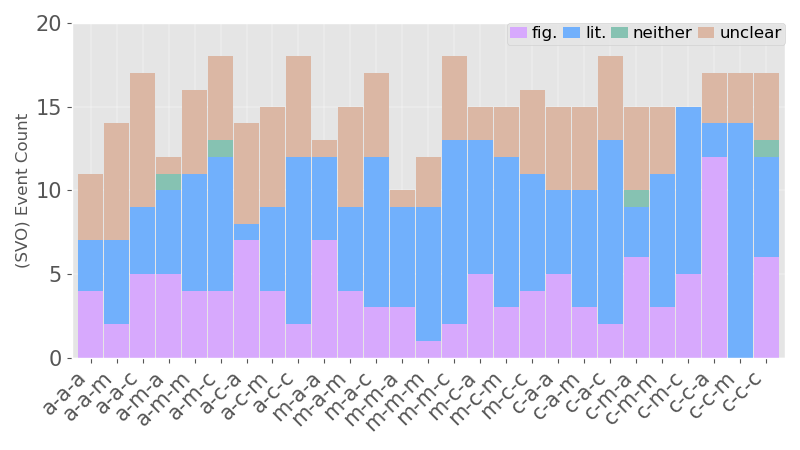}
        \caption{\textbf{\texttt{Llama3.1-8B}} judgements across abstract. comb.}
    \end{subfigure}%
         \begin{subfigure}[b]{0.225\textwidth}
         \centering
        \includegraphics[width=\linewidth]{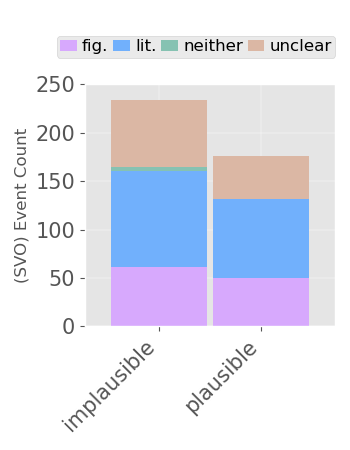}
        \caption{orig. label}
    \end{subfigure}%
    \begin{subfigure}[b]{0.225\textwidth}
        \centering
        \includegraphics[width=\linewidth]{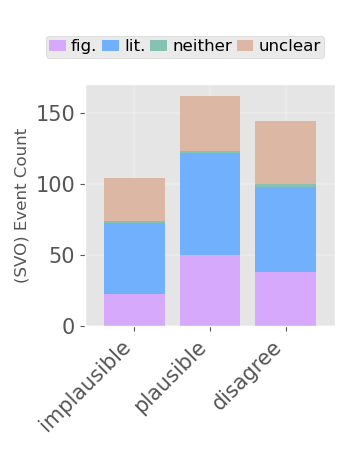}
        \caption{PAP rating}
    \end{subfigure}
    \hfill
    \setcounter{subfigure}{0}
\renewcommand{\thesubfigure}{4\alph{subfigure}}
    \begin{subfigure}[b]{0.5\textwidth}
         \centering
        \includegraphics[width=\linewidth]{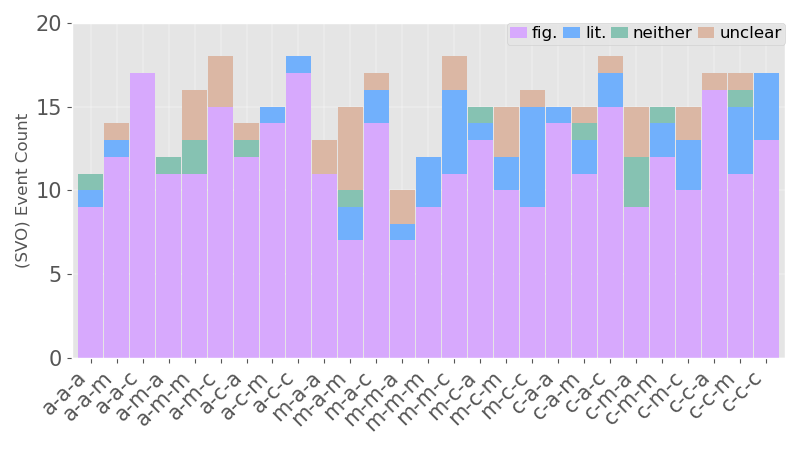}
        \caption{\textbf{\texttt{Gemma3-4B}} judgements across abstract. comb.}
    \end{subfigure}%
         \begin{subfigure}[b]{0.225\textwidth}
         \centering
        \includegraphics[width=\linewidth]{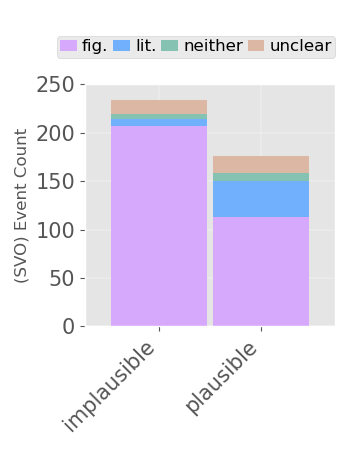}
        \caption{orig. label}
    \end{subfigure}%
    \begin{subfigure}[b]{0.225\textwidth}
        \centering
        \includegraphics[width=\linewidth]{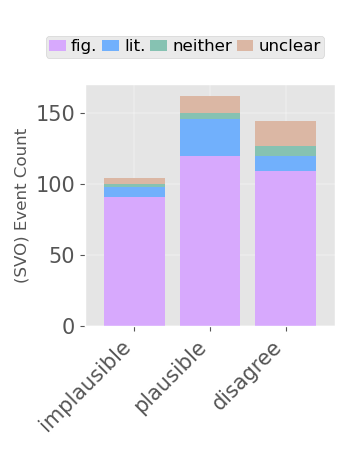}
        \caption{PAP rating}
    \end{subfigure}
    \hfill
    \caption{Overview of analysis of \textbf{\texttt{Qwen3-4B}},  \textbf{\texttt{Mistral-7B}},  \textbf{\texttt{Llama3.1-8B}}, and \textbf{\texttt{Gemma3-4B}} figurative labels (\textbf{zero-shot}), similarly to human analysis in Figure~\ref{fig:human-label-analysis}.}
    \label{fig:full-model-label-analysis}
    \vspace{-0.3em}
\end{figure*}

\begin{figure*}[!htbp]
    \centering
        \setcounter{subfigure}{0}
    \renewcommand{\thesubfigure}{1\alph{subfigure}}
    \begin{subfigure}[b]{0.5\textwidth}
         \centering
        \includegraphics[width=\linewidth]{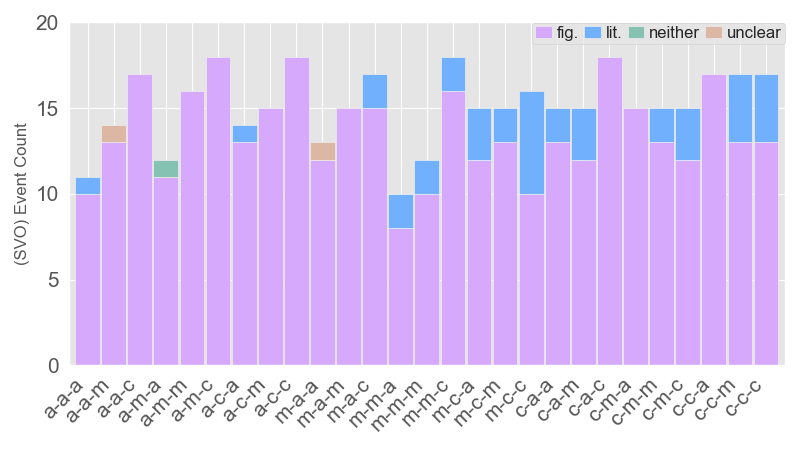}
        \caption{\textbf{\texttt{Qwen3-4B}} judgements across abstractness comb.}
    \end{subfigure}%
         \begin{subfigure}[b]{0.225\textwidth}
         \centering
        \includegraphics[width=\linewidth]{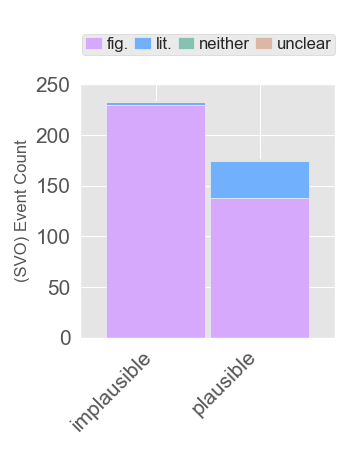}
        \caption{orig. label}
    \end{subfigure}%
    \begin{subfigure}[b]{0.225\textwidth}
        \centering
        \includegraphics[width=\linewidth]{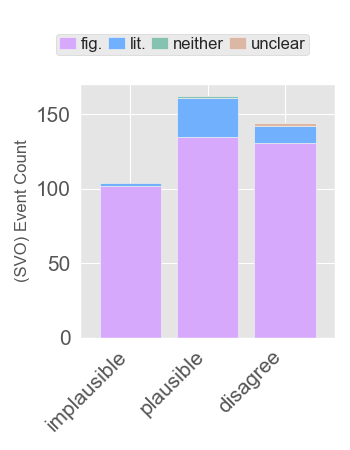}
        \caption{PAP rating}
    \end{subfigure}
    \hfill
    \setcounter{subfigure}{0}
    \renewcommand{\thesubfigure}{2\alph{subfigure}}
    \begin{subfigure}[b]{0.5\textwidth}
         \centering
        \includegraphics[width=\linewidth]{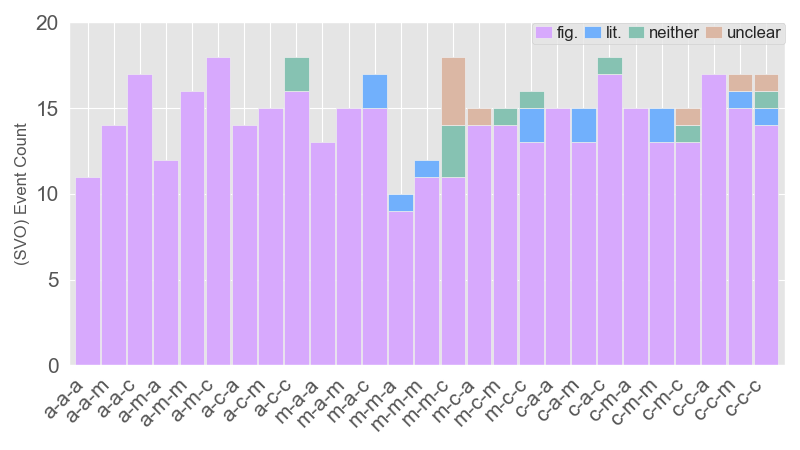}
        \caption{\textbf{\texttt{Mistral-7B}} judgements across abstractness comb.}
    \end{subfigure}%
         \begin{subfigure}[b]{0.225\textwidth}
         \centering
        \includegraphics[width=\linewidth]{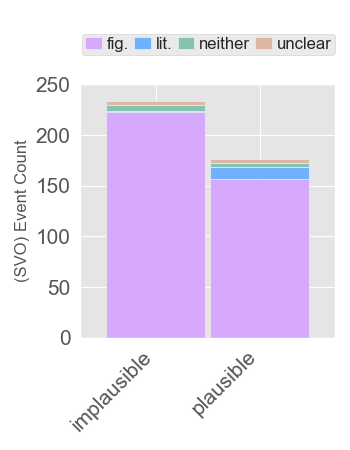}
        \caption{orig. label}
    \end{subfigure}%
    \begin{subfigure}[b]{0.225\textwidth}
        \centering
        \includegraphics[width=\linewidth]{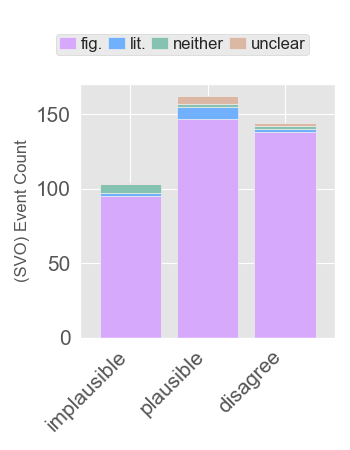}
        \caption{PAP rating}
    \end{subfigure}
    \hfill
\setcounter{subfigure}{0}
\renewcommand{\thesubfigure}{3\alph{subfigure}}
    \begin{subfigure}[b]{0.5\textwidth}
         \centering
        \includegraphics[width=\linewidth]{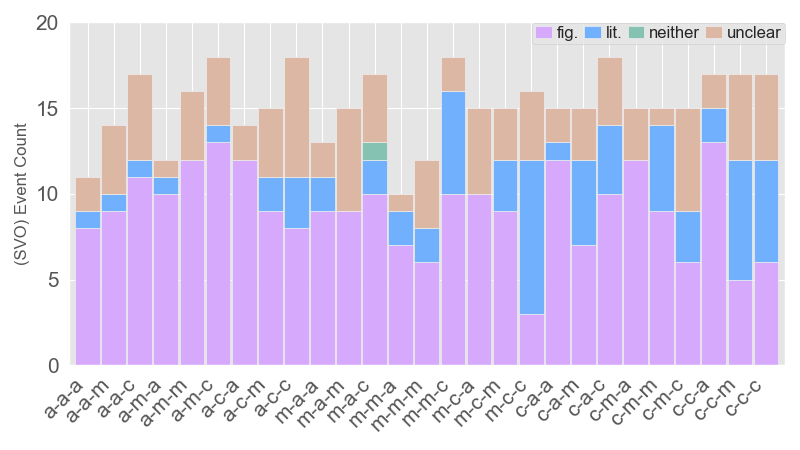}
        \caption{\textbf{\texttt{Llama3.1-8B}} judgements across abstract. comb.}
    \end{subfigure}%
         \begin{subfigure}[b]{0.225\textwidth}
         \centering
        \includegraphics[width=\linewidth]{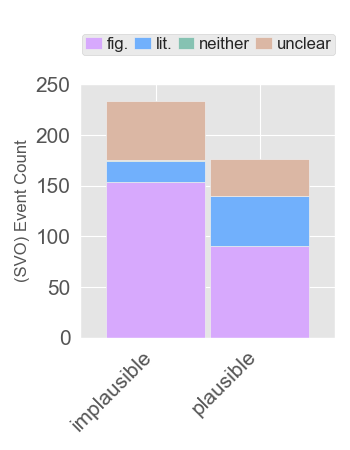}
        \caption{orig. label}
    \end{subfigure}%
    \begin{subfigure}[b]{0.225\textwidth}
        \centering
        \includegraphics[width=\linewidth]{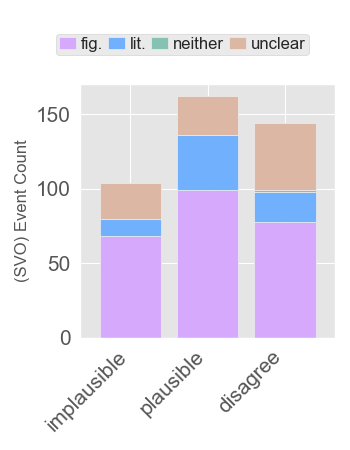}
        \caption{PAP rating}
    \end{subfigure}
    \hfill
    \setcounter{subfigure}{0}
\renewcommand{\thesubfigure}{4\alph{subfigure}}
    \begin{subfigure}[b]{0.5\textwidth}
         \centering
        \includegraphics[width=\linewidth]{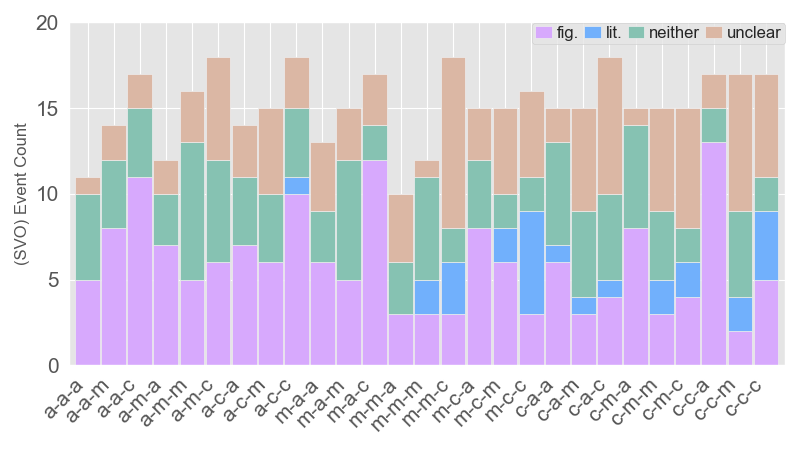}
        \caption{\textbf{\texttt{Gemma3-4B}} judgements across abstract. comb.}
    \end{subfigure}%
         \begin{subfigure}[b]{0.225\textwidth}
         \centering
        \includegraphics[width=\linewidth]{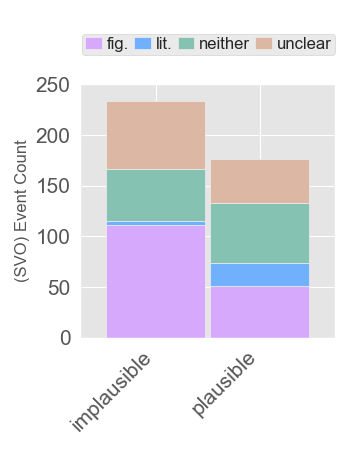}
        \caption{orig. label}
    \end{subfigure}%
    \begin{subfigure}[b]{0.225\textwidth}
        \centering
        \includegraphics[width=\linewidth]{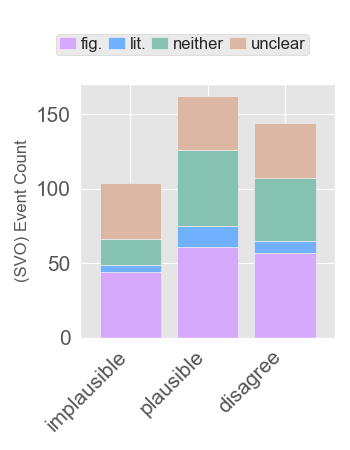}
        \caption{PAP rating}
    \end{subfigure}%
    \hfill
    \caption{Overview of analysis of \textbf{\texttt{Qwen3-4B}},  \textbf{\texttt{Mistral-7B}},  \textbf{\texttt{Llama3.1-8B}}, and \textbf{\texttt{Gemma3-4B}} figurative labels (\textbf{few-shot}), similarly to human analysis in Figure~\ref{fig:human-label-analysis}.}
    \label{fig:full-model-label-analysis-fewshot}
    \vspace{-0.3em}
\end{figure*}

We present \textbf{zero-shot} model results for predicting figurative labels for all models in Figure~\ref{fig:full-model-label-analysis}. \textbf{Few-shot} model results for predicting figurative labels are presented for all models in Figure~\ref{fig:full-model-label-analysis-fewshot}. 

Further, quantitative analysis results for few-shot modeling results are presented in Table~\ref{tab:label-specific-chi-cramersv-all-fewshot}. We include values from human analysis for reference. 

We analyze which prompt template leads to strongest bias towards figurative interpretation. We consider both zero- and few-shot prompt templates and show results in Table~\ref{tab:prompt-analysis}.



\begin{figure*}[!htpb]
\vspace{-0.2cm}
    \centering

\begin{lstlisting}[language=Python, caption={Question}]
Is the following event figurative or literal or neither?
Event: {event}
Answer with only one label: figurative or literal or neither.
Respond in the following format:
Label: <figurative|literal|neither>
\end{lstlisting}
\begin{lstlisting}[language=Python, caption={Statement}]
Determine whether the event below is figurative or literal or neither.
Event: {event}
Respond in the following format:
Label: <figurative|literal|neither>
\end{lstlisting}
\begin{lstlisting}[language=Python, caption={Human instruction-based}]
Decide whether the event description is clearly based on the meanings of the three words.
Event: {event}
Answer with only one label: <figurative|literal|neither>.
Decision: <label>
\end{lstlisting}
    \caption{\textbf{Zero shot} prompt templates, from top to bottom including the task formulation as a \textit{question} (top), a \textit{statement }(middle), and as closely as possible based on the \textit{instructions for human annotation} but condensed in one sentence (bottom).}
    \label{fig:prompts-few-shot}
\end{figure*}

\begin{figure*}[!htpb]
\vspace{-0.2cm}
    \centering
\begin{lstlisting}[language=Python, caption={Human instruction-based}]
Decide whether the event description is clearly based on the meanings of the three words.
For example, the event "cat eat sardine" is literally describing the event of a cat eating a sardine.
In contrast, the event "friend grasp meaning" does not describe a friend literally grasping a meaning:
the event meaning is figurative, rather than literal.
Event: {event}
Answer with only one label: <figurative|literal|neither>
Decision: <label>
\end{lstlisting}

    \caption{\textbf{Few shot} prompt template. The prompt is based as closely as possible on the \textit{instructions for human annotation} while optimizing for the shortest length.}
    \label{fig:prompts-zero-shot}
\end{figure*}

\setlength{\abovecaptionskip}{2pt}
\setlength{\belowcaptionskip}{2pt}

\begin{table*}[!htpb]
\centering
\small
\begin{tabular}{@{}l|r|rr|rr|rr|rr|rr@{}}
\toprule

\multicolumn{2}{l}{\hspace*{-2mm}\textbf{}}
& \multicolumn{2}{l}{\textbf{Human}} & \multicolumn{2}{l}{\textbf{Qwen3-4B}} & \multicolumn{2}{l}{\textbf{Llama3.1-8B}} & \multicolumn{2}{l}{\textbf{Mistral-7B}} & \multicolumn{2}{l}{\textbf{Gemma3-4B}} \\ \midrule
\textbf{Figurative}  &  df & $\chi^{2}$    & $V$ & $\chi^{2}$    & $V$ & $\chi^{2}$    & $V$ & $\chi^{2}$    & $V$ & $\chi^{2}$    & $V$           \\
\midrule

\textsc{abstractness} & 26 & \textbf{63.64***}  & 0.39 & 43.66*                 & 0.33               & 46.71**  & 0.34 & 51.37** & 0.35 & 53.59**            & 0.36          \\
\textsc{orig. label}       & 1  & 6.91**    & 0.13 & 41.05                  & 0.32               & 7.74**   & 0.14 & 4.64*   & 0.11 & \textbf{13.56***}           & 0.18          \\
\textsc{PAP\_rating} & 3  & 13.49**   & 0.18 & 15.48**                & 0.19               & 4.04     & 0.10  & 3.37    & 0.09 & 1.23               & 0.05          \\\midrule
\multicolumn{12}{l}{\hspace*{-2mm}\textbf{Literal}}\\
\midrule
\textsc{abstractness} & 26 & 43.93**   & 0.33 & 50.84**                & 0.35               & \textbf{61.28***} & 0.39 & 34.08   & 0.29 & \textbf{57.32***}          & 0.37          \\
\textsc{orig. label}       & 1  & \textbf{111.33***} & 0.52 & 40.70                   & 0.32               & \textbf{25.35***} & 0.25 & 10.02** & 0.16 & \textbf{19.26***}           & 0.22          \\
\textsc{pap\_rating} & 3  & \textbf{17.29*** } & 0.21 & 15.73**                & 0.2                & 7.37     & 0.13 & 3.91    & 0.10  & 1.97               & 0.07          \\ \midrule
\multicolumn{12}{l}{\hspace*{-2mm}\textbf{Neither}}\\
\midrule
\textsc{abstractness} & 26 & 32.80      & 0.28 & 33.25                  & 0.28               & 23.17    & 0.24 & 33.14   & 0.28 & 29.15              & 0.27          \\
\textsc{orig. label}       & 1  & 71.96     & 0.42 & 0.02                   & 0.01               & 0        & 0    & -       & -    & 6.45*              & 0.13          \\
\textsc{pap\_rating} & 3  & 13.73**   & 0.18 & 1.54                   & 0.06               & 1.86     & 0.07 & 6.54    & 0.13 & 8.39*              & 0.14          \\ \midrule
\multicolumn{12}{l}{\hspace*{-2mm}\textbf{Unclear}}\\
\midrule
\textsc{abstractness} & 26 & 29.59     & 0.27 & 28.55                  & 0.26               & 20.12    & 0.22 & 51.42** & 0.35 & 39.34*             & 0.31          \\ 
\textsc{orig. label}       & 1  & 8.23**    & 0.14 & 0                      & 0                  & 1.02     & 0.05 & 0       & 0    & 0.87               & 0.05          \\
\textsc{pap\_rating} & 3  & 4.17      & 0.10  & 3.73                   & 0.10                & 10.21*   & 0.16 & 1.89    & 0.07 & 9.44**             & 0.15        \\ \bottomrule

\end{tabular}
\caption{Associations between figurative language and abstractness, original label, or PAP ratings. $\chi^{2}$ indicates \textit{significance} ($p<0.05$:*, $p<0.01$:**, $p<0.001$:***) and Cramér's $V$ measures \textit{strength} of association. Model results are based on \textbf{few-shot} prompts.} 
\label{tab:label-specific-chi-cramersv-all-fewshot}
\vspace{-1.7em}
\end{table*}

\begin{table*}[!htpb]
\centering
\begin{tabular}{@{}l|rrr|r@{}}
\toprule
  \multicolumn{1}{c}{}       & \multicolumn{3}{c}{\textsc{\textbf{Zero-shot}}}                                                    & \multicolumn{1}{c}{\textsc{\textbf{Few-shot}}} \\ \cmidrule{2-5}
  \multicolumn{1}{c}{}       & \multicolumn{1}{c}{question} & \multicolumn{1}{c}{statement} & \multicolumn{1}{c}{human instr.} & \multicolumn{1}{c}{human instr.}    \\ \midrule
Gemma3-4B   & 67.88                      & 89.29                           & 61.05                     & 22.83                        \\
Qwen3-4B    & 91.00                      & 83.94                           & 63.50                      & 73.90                        \\
Mistral-7B & 84.91                      & 95.83                           & 78.47                     & 94.51                        \\
Llama3.1-8B   & 23.88                      & 31.39                           & 31.03                     & 61.37         \\ \bottomrule             
\end{tabular}
\caption{Overview of share of \underline{only} figurative labels per prompt, in percent. Note that percentages do not add up to 100\% but each number is a share of 100\% labels distributed among \textit{figurative} (shown here), \textit{literal}, \textit{neither}, and \textit{unclear} (not shown). For reference, prompt templates are listed in Figure~\ref{fig:prompts-zero-shot} and Figure~\ref{fig:prompts-few-shot}.}
\label{tab:prompt-analysis}
\end{table*}


\end{appendices}

\end{document}